\RequirePackage[loading]{tracefnt}
\documentclass{article}
\usepackage[weather]{ifsym}
\usepackage{ijcai19}

\usepackage{times}
\usepackage{soul}
\usepackage{url}
\usepackage[hidelinks]{hyperref}
\usepackage[utf8]{inputenc}
\usepackage[small]{caption}
\usepackage{graphicx}
\usepackage{amsmath}
\usepackage{booktabs}
\usepackage{subfigure,tabularx,algorithmic,units,hyperref}
\usepackage{algorithm}
\usepackage{amsfonts}
\usepackage{dsfont}
\usepackage{graphicx}
\usepackage{amsthm}
\newtheorem{definition}{Definition}

\title{The Dangers of Post-hoc Interpretability: Unjustified Counterfactual Explanations}

\author{
Thibault Laugel$^1$\footnote{Contact Author}\and
Marie-Jeanne Lesot$^1$\and
Christophe Marsala$^{1}$\and\\
Xavier Renard$^2$\And
Marcin Detyniecki$^{1,2,3}$\\
\affiliations
$^1$Sorbonne Université, CNRS, Laboratoire d'Informatique de Paris 6, LIP6, F-75005 Paris, France\\
$^2$AXA, Paris, France\\
$^3$Polish Academy of Science, IBS PAN, Warsaw, Poland\\
\emails
thibault.laugel@lip6.fr
}

\begin{document}

\maketitle

\begin{abstract}
Post-hoc interpretability approaches have been proven to be powerful tools to generate explanations for the predictions made by a trained black-box model. However, they create the risk of having explanations that are a result of some artifacts learned by the model instead of actual knowledge from the data. This paper focuses on the case of counterfactual explanations and asks whether the generated instances can be \textit{justified}, i.e. continuously connected to some ground-truth data. We evaluate the risk of generating unjustified counterfactual examples by investigating the local neighborhoods of instances whose predictions are to be explained and show that this risk is quite high for several datasets.
Furthermore, we show that most state of the art approaches do not differentiate justified from unjustified counterfactual examples, leading to less useful explanations.

\end{abstract}

\section{Introduction}
Among the soaring number of methods proposed to generate explanations for a machine learning classifier, post-hoc interpretability aproaches~\cite{Guidotti2018survey} have been the subject of debates recently in the community~\cite{Rudin2018}. By generating explanations for the predictions of a trained predictive model without using any knowledge about it whatsoever (treating it as a \textit{black-box}), these systems are inherently flexible enough to be used in any situation~(model, task...) by any user, which makes them popular today in various industries. However, their main downside is that, under these assumptions, there is no guarantee that the built explanations  are faithful to the original data that were used to train the model.

This is in particular the case for counterfactual example approaches that, based on counterfactual reasoning (see e.g. \cite{Bottou2013}), aim at answering the question: \textit{given a classifier and an observation, how is the prediction altered when the observation changes?}
In the context of classification, they identify the minimal perturbation required to change the predicted class of a given observation: a user is thus able to understand what features locally impact the prediction and therefore how it can be changed.
Among interpretability methods, counterfactual examples have been shown to be useful solutions~\cite{Wachter2018} that can be easily understood and thus directly facilitate decisions for a user.

However, without any guarantee on existing data, counterfactual examples in the post-hoc paradigm are vulnerable to issues raised by the robustness of the classifier, leading to explanations that are arguably not satisfying in the context of interpretability.
More generally, this paper argues that a crucial property a counterfactual example should satisfy is that it should be connected to the training data of the classifier. Formally, we define this relation using the notion of connectedness and argue that there should be a continuous path between a counterfactual and an instance from the training data, as discussed and defined in Section~\ref{sec:justification}.

This paper aims at showing that generating such post-hoc counterfactual explanations can be difficult. The contributions are the following:

\begin{itemize}
    \item We propose an intuitive desideratum for more relevant counterfactual explanations, based on ground-truth labelled data, that helps generating better explanations.
    \item We design a test to highlight the risk of having undesirable counterfactual examples disturb the generation of counterfactual explanations.
    \item We apply this test to several datasets and classifiers and show that the risk of generating undesirable counterfactual examples is high.
    \item Additionally, we design a second test and show that state of the art post-hoc counterfactual approaches may generate unjustified explanations.
\end{itemize}

Section~\ref{sec:background} is devoted to discussing the state of the art of post-hoc interpretability and counterfactual explanations, as well as presenting studies related to this work. Section~\ref{sec:justification} proposes a definition for ground-truth backed counterfactual explanations that we use in this paper. In Section~\ref{sec:detection}, the central procedure of this paper is proposed, which aims at assessing the risk of generating unjustified counterfactual explanations, and experimental results are shown. Finally, the vulnerability of methods from the literature to this risk is evaluated in Section~\ref{sec:evaluation}.

\section{Background}
\label{sec:background}

This section presents some existing interpretability methods and analyses that have been already conducted.

\subsection{Post-hoc Interpretability}

In order to build explanations for predictions made by a trained black-box model, post-hoc interpretability approaches generally rely on sampling instances and labelling them with the model to gather information about its behavior, either locally around a prediction~\cite{Ribeiro2016} or globally~\cite{Craven1996}. These instances are then used to approximate the decision function of the black-box and build understandable explanations, either using a surrogate model (e.g. a linear model~\cite{Ribeiro2016} or a decision tree~\cite{Hara2016}), or by computing meaningful coefficients~\cite{Turner2015,Lundberg2017}). Other methods rely on using specific instances to build explanations using comparison to relevant neighbors, such as prototype-based approaches~\cite{Kim2014} and counterfactual explanation approaches. 

Instead of simply identifying important features, the latter aim at finding the minimal perturbation required to alter the prediction of a given observation. A counterfactual explanation is thus a data instance, close to this observation, that is predicted to belong to a different class.
This form of explanation provides a user tangible explanations that are directly understandable and actionable, 
which can be seen as an advantage.
This can be opposed to other explanations using feature importance vectors, which are arguably harder to use and to understand for a non-expert user~\cite{Wachter2018}.
Several formalizations of the counterfactual problem can be found in the literature, depending on the formulation of the inverse problem and on the used distance.
For instance,~\cite{Laugel2017inverse} (resp.~\cite{Guidotti2018lore}) look for the $L_2$ (resp. $L_0$)-closest instance of an other class, while~\cite{Lash2017} try to find the instance that has the highest probability of belonging to another class within a certain maximum distance. Another example is the formulation as a tradeoff between the $L_1$ closest instance and a certain classification score target~\cite{Wachter2018}.

\subsection{Comparison of Post-hoc Interpretability Approaches}

The post-hoc paradigm, and in particular the need for post-hoc approaches to use instances that were not used to train the model to build their explanations, raises questions about their relevance and usefulness.
Several troublesome issues have been identified:~\cite{Baehrens2010} e.g. notice that modeling the decision function of a black-box classifier with a surrogate model trained on generated instances can result in explanation vectors that point in the wrong directions in some areas of the feature space in a trivial problem.
The stability of post-hoc explainer systems has been criticized as well, showing that some of these approaches are locally not stable enough~\cite{Alvarez2018} or on the contrary too stable and thus not locally accurate enough~\cite{Laugel2018}.

However, there seems to have been few works~\cite{Lipton2017,Rudin2018} questioning the post-hoc paradigm itself, and wondering about the risks of generating explanations that are disconnected from the ground truth. The work of \cite{Kabra2015} can be related to this issue, as it attempts to identify which neighbors from the training data are the most influential for a given prediction made by a trained classifier. However, their approach relies on retraining the classifier to induce a notion of causality, and thus lies beyond the post-hoc context.
In a similar fashion, \cite{Jiang2018} try to identify which predictions can be trusted in regard to ground truth instances based on their distance to training data.

\section{Ground-truth Justification}
\label{sec:justification}

This section presents motivations and formalization for the notion of ground-truth justification, central to this paper.

\begin{figure}[t]
    \centering
        \subfigure[RF with 3 estimators]{\includegraphics[width=0.48\columnwidth]{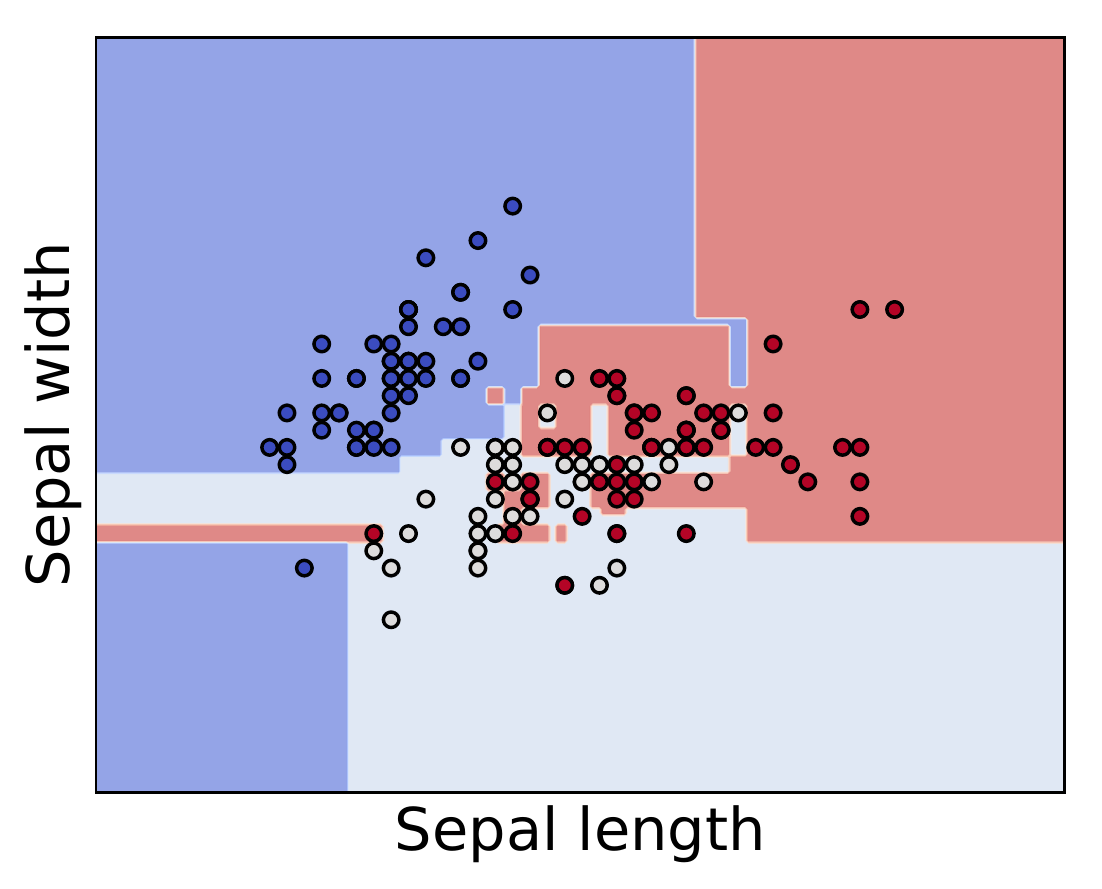}}
        \subfigure[SVC with RBF kernel]{\includegraphics[width=0.48\columnwidth]{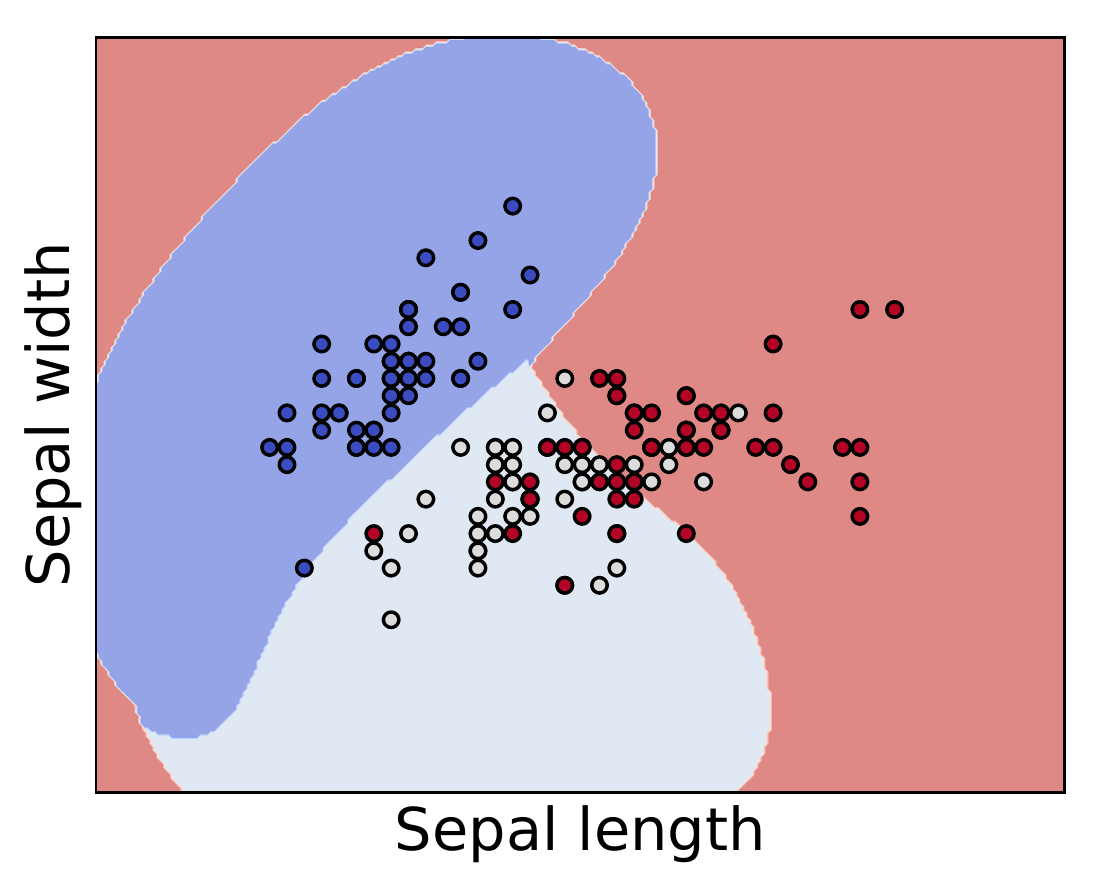}}
    \caption{Two classifiers have been trained on $80\%$ of the dataset (a 2D version of the iris dataset) and have the same accuracy over the test set, $0.78$. Left picture: because of its low robustness, the random forest classifier makes questionable generalizations (e.g. small red square in the dark blue region)
    Right picture: the support vector classifier makes questionable decisions in regions far away from the training data (red area in the top left corner).}
    \label{fig:iris-illustration}
    \vspace{-0.3cm}
\end{figure}

\subsection{Motivations}

The notion of ground-truth justification we seek aims at making a distinction between an explanation that has been generated because of some previous knowledge (such as training data) and one that would be a consequence of an artifact of the classifier. In the post-hoc paradigm, explainer systems do not have access to the training data; making this distinction is thus not trivial.
In the case of counterfactual explanations that rely on specific instances, the risk would be having counterfactual instances (i.e. predictions from another class) caused by these artifacts.
Artifacts can be created in particular because of a lack of robustness of the model, or because it is forced to make a prediction for an observation in an area that was not being covered by the training set. 
Fig.~\ref{fig:iris-illustration} shows an illustration of how easily such situations can arise in a trivial problem. In both cases, some regions can be found where the classifier makes questionable improvisations in areas he has no information about (no training data), either because it is not robust enough (left image) or because of its complexity (right image). 
Although not harmful in the context of prediction (a desirable property of a classifier remains its ability to generalize to new observations), having an explanation caused by an artifact such that it can not be associated to any existing knowledge by a human user may be undesirable in the context of interpretability: for instance, a physician using a diagnostic tool and an explanation for a prediction that is not based on existing medical cases would be conceptually useless, if not very dangerous.

\subsection{Definitions}

We propose to define this relation between an explanation and some existing knowledge (ground-truth data used to train the black-box model) using the topological notion of path, used when defining the path connectedness of a set. In order to be more easily understood and employed by a user, we argue that the counterfactual instance should be continuously connected to an observation from the training dataset. The idea of this \textit{justification} is not to identify the instances that are \textit{responsible} for a prediction (such as in the aforementioned work of~\cite{Kabra2015}), but the ones that are correctly being predicted to belong to the same class for similar reasons.

\begin{definition}[Justification]
Given a classifier $f:\mathcal{X}\rightarrow\mathcal{Y}$ trained on a dataset $X$, a counterfactual example $e \in \mathcal{X}$ is \emph{justified} by an instance $a \in X$ correctly predicted if $f(e) = f(a)$ and if there exists a continuous path $h$ between $e$ and $a$ such that no decision boundary of $f$ is met.

Formally, $e$ is \emph{justified} by $a \in X  \; \text{if:} \; \exists \; h:[0, 1] \rightarrow \mathcal{X}$ such that: (i) $h$ is continuous, (ii) $h(0) = a$, (iii) $h(1)=e$ and (iv) $\forall{t}\in [0,1],\,f(h(t))=f(e)$.

\label{def:justification}
\end{definition}

To adapt this continuous notion to the context of a black-box classifier, we replace the connectedness constraint with \textit{$\epsilon$-chainability}, with $\epsilon \in \mathbb{R}^+$: an \emph{$\epsilon$-chain} between $e$ and $a$ is a finite sequence $e_0$, $e_1$, ... $e_N$ $\in \mathcal{X}$ such that $e_0=e$, $e_N=a$ and $\forall{i} < N,\, d(e_i, e_{i+1})<\epsilon$, with $d$ a distance metric.

\begin{definition}[$\epsilon$-justification]
Given a trained classifier $f:\mathcal{X}\rightarrow\mathcal{Y}$ trained on a dataset $X$, a counterfactual example $e \in \mathcal{X}$ is \emph{$\epsilon$-justified} by an instance $a \in X$ correctly predicted if $f(e) = f(a)$ and if there exists an $\epsilon$-chain $\{e_i\}_{i\leq N}$ between $e$ and $a$ such that $\forall i \leq N, f(e_i)=f(e)$.
\label{def:epsilon-justification}
\end{definition}

This definition is equivalent to approximating the aforementioned function $h$ by a sequence $(e_i)_i$:, when $\epsilon$ decreases towards 0, this definition becomes a weak version of Definition~\ref{def:justification}.
Consequently, we call a \emph{justified} (resp. \emph{unjustified}) counterfactual (written JCF, resp. UCF) a counterfactual example that does (resp. does not) satisfy Definition~\ref{def:epsilon-justification}. 

Setting parameter $\epsilon$ then allows to build an $\epsilon$-graph of instances classified similarly. Using such a graph to approximate connectedness is also found in some manifold learning approaches~(e.g. Isomap~\cite{Tenenbaum2000}), where local neighborhoods help to approximate connectedness and thus reduce the dimension of the data.

This definition of connectedness can also be related to the one used in density-based clustering methods, such as the well-known DBSCAN~\cite{Ester96}: in DBSCAN, two parameters $\epsilon$ and $minPts$ control the resulting clusters (resp. outliers), built from the core points, which are instances that have at least (resp. less than) $minPts$ neighbors in an $\epsilon$-ball. Thus, having an instance being $\epsilon$-justified by another is equivalent to having them both belong to the same DBSCAN cluster, with parameters $minPts=2$ and same $\epsilon$.

\section{Assessment of the Local Risk of Generating Unjustified Counterfactuals}
\label{sec:detection}

The goal of this section is to design a test, the \textit{Local Risk Assessment} (LRA) procedure, highlighting the existence of UCF in low-dimensional structured data.

\subsection{LRA Procedure}
\label{sec:detection-procedure}

\begin{figure}[t]
\centering
\subfigure{\includegraphics[width=0.47\columnwidth,height=2.8cm]
{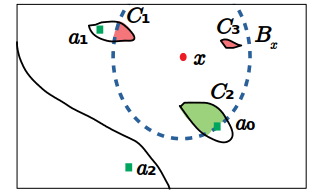}
}
\medskip
\subfigure{\includegraphics[width=0.47\columnwidth,height=2.8cm]{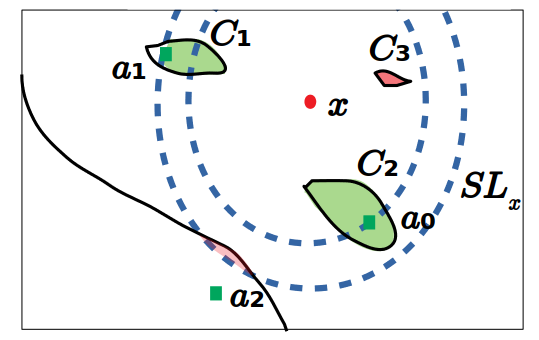}
}
\vspace{-0.5cm}
\caption{Illustration of the Local Risk Assessment procedure in the context of binary classification. Left: Definition and Initial Assessment steps; right: Iteration step.}
\label{fig:illustration-algo}
\end{figure}

Given a black-box classifier $f:\mathcal{X}\rightarrow \mathcal{Y}$ trained on the dataset~$X$ of instances of~$\mathcal{X}$ and an instance $x\in \mathcal{X}$ whose prediction $f(x)$ is to be explained, the aim is to assess the risk of generating unjustified counterfactual examples in a local neighborhood of $x$. 
To do this, we propose a generative approach that aims at finding which regions of this neighborhood are $\epsilon$-connected to an instance of $X$ (i.e. verify Definition~\ref{def:epsilon-justification}).
In the rest of the paper, given a subset of instances $A$ of $\mathcal{X}$, we note $A^l = \{z \in A~|~f(z)=l\}$.
The LRA procedure, commented below, is detailed in Algorithm~\ref{alg:lra} and illustrated in a two-dimensional setting in  Fig.~\ref{fig:illustration-algo}: the red dot represents $x$, the observation whose prediction is to be interpreted, while green squares ($a_0$, $a_1$ and $a_2$) are correctly classified instances from the training set $X$. The decision boundary of $f$, the considered binary classifier, is represented by the black lines. 

For clarity purposes, the procedure is split in three steps: first, the studied area is defined in the Definition step. Then, an Initial Risk Assessment step is performed in this area. Finally, if needed, the procedure is repeated in the Iteration step.

\subsubsection{Definition Step}
We first define the studied local neighborhood, i.e. the region of $x$ which we are trying to assess the risk of, as the ball with center $x$ and radius defined by its distance to its closest neighbor from $X$ correctly predicted to belong to another class:  $\mathcal{B}(x, d(x, a_0))$, with ${a_0}~=~\underset{z \in X^{l\neq f(x)}}{\text{arg min}} d(x, z), \text{ s.t. } f(a_0) \text{ is correct}$.
The distance $d(x, a_0)$ represents an approximation of the minimal distance between $x$ and the decision boundary of $f$, and is hence a reasonable distance to look for counterfactual examples: as it is $\epsilon$-connected to itself, $a_0$ represents a "close" justified counterfactual explanation by itself. The limit of $\mathcal{B}(x, d(x, a_0))$ is illustrated in the left picture of Fig.~\ref{fig:illustration-algo} by the blue dashed circle. 

\subsubsection{Initial Assessment Step}
A high number $n$ of  instances $B_x = \{x_i\}_{i\leq n}$ are then sampled in $\mathcal{B}(x, d(x, a_0))$ following a uniform distribution and labelled using $f$.

Using Definition~\ref{def:epsilon-justification}, the goal of LRA is to identify which instances of $B_x^{f(a_0)}$ are connected to $X$ through an $\epsilon$-chain. The process to set the value of $\epsilon$ is detailed in Section~\ref{sec:epsilon}.
As mentioned earlier, an easy way to implement this is to use the clustering algorithm DBSCAN on $B_x^{f(a_0)} \cup \{a_0\}$ with parameter values $\epsilon$ and $minPts = 2$.
We note $\{C_t\}_t$ 
the resulting clusters and outliers: according to Def.~\ref{def:epsilon-justification}, the instances belonging to the same cluster as $a_0$ are $\epsilon$-justified. 

This Initial Assessment step is written in lines 2 to 9 of Algorithm~\ref{alg:lra}, and illustrated in the left image of Fig.~\ref{fig:illustration-algo}: instances are generated in $B_x$ (blue dashed circle; the generated instances are not shown in the illustration), and the ones belonging to $B_x^{f(a_0)}$ are assigned to clusters $C_1$ (unjustified at this step), $C_2$ (justified, since $a_0 \in C_2$) and $C_3$ (non-justified).

\subsubsection{Iteration Step}
However, at this step, there is no certainty that the other instances of~$B^{f(a_0)}$  are unjustified, as they could simply be connected to other instances from~$X^{f(a_0)}$ than~$a_0$, or using an $\epsilon$-path that can not be fully included in the explored area~$\mathcal{B}(x, d(x, a_0))$ (see Fig.~\ref{fig:illustration-algo}, instances belonging to cluster~$C_1$ are actually justified by $a_1$ but are not connected within $B_x$).
To answer this question, we define $a_1$ as the second closest instance of $X^{f(a_0)}$ to $x$ correctly predicted and broaden the exploration region to the hyperspherical layer defined as:
$$ \mathcal{SL}_1 =  \{x\in \mathcal{X} \text{ s.t. } d(a_0, x_0)\leq d(x,x_0) \leq d(a_1, x_0)\}.$$
Instances $SL_1$ are generated uniformly in~$\mathcal{SL}_1$, and the instances from~$SL_1^{f(a_0)} \cup \{a_1\}$ can then be assigned either to previous clusters~$C_t$ or to $a_1$ using the same criteria, i.e. if their minimum distance to an instance of an existing cluster is less than $\epsilon$ (right image of Fig.~\ref{fig:illustration-algo}: cluster~$C_1$ can now be connected to $a_1$ through the instances generated in $SL_1$).

This step is repeated by generating hyperspherical layers defined by all instances from~$X^{f(a_0)}$ until all the initial clusters~$C_t$ can either be justified or are not being updated by any new instances anymore (such as cluster~$C_3$ of Fig.~\ref{fig:illustration-algo}, which was not updated between the two iterations). This step is illustrated in the "while" loop of Algorithm~\ref{alg:lra}, lines~10 to~19). If some non-connected clusters are still being updated when all instances from the training set have been explored (e.g. red region in the top left corner of the left picture of Fig.~\ref{fig:iris-illustration}), they are considered as non-connected.

In the end, the LRA procedure returns~$n_J$ (resp.~$n_U$) the number of JCF (resp. UCF) originally generated in~$\mathcal{B}(x, d(x, a_0))$. If $n_U > 0$, there exists a risk of generating unjustified counterfactual examples in the studied area.

\begin{algorithm}[t!]
\caption{Local risk assessment}
\label{alg:lra}
\begin{algorithmic}[1]
 \REQUIRE $x$, $f$, $X$
 \STATE Sort correctly predicted instances from $X^{l \neq f(x)}=\{a_0, a_1, ...\}$ in increasing order of their distance to $x$ \;
 
 \STATE $B_x = \{x_i\}_{i\leq n} \sim \text{Uniform}(\mathcal{B}(x, a_0)$) \;
 
 \STATE $\epsilon = \underset{x_i \in B_x}{\text{max }}\underset{x_j \in B_x}{\text{min }} d(x_i, x_j)$ \;
 
 \STATE $B_x^{f(a_0)} = \{x_i \in B_x\,:\,f(x_i)=f(a_0)\} $ \;
 
 \STATE $\{C_t\}_t \leftarrow \text{DBSCAN} ( B_x^{f(a_0)} \cup \{a_0\}, \epsilon, \text{minPts}=2)$ \;
 
 \STATE $\mathcal{C}_J = \{C_{t_0}\} \,\,\,\text{s.t.}\,\, a_0 \in  C_{t_0} \,\,\,;\,\,\, n_J = |\mathcal{C}_J| $\;
 
 \STATE $\mathcal{C}_{NC} = \{C_t\}_t \setminus C_{t_0} \,\,\,;\,\,\, n_{NC} = |\mathcal{C}_{NC}|$\;
 
 \STATE $\mathcal{C}_{U} = \{\}  \,\,\,;\,\,\, n_{U} = 0 $\;
 \STATE $ k = 0$ \;

 \WHILE{$ n_{NC} > 0$}
 
 \STATE $k = k + 1$ \;
 
 \STATE $SL_k = \{x_i\}_i \sim \text{Uniform}(\mathcal{SL}_k)$ \;
  
  \STATE $SL_k^{f(a_k)} = \{x_i \in SL_k\,:\,f(x_i)=f(a_k)\}$ \;
  
  \STATE $\{C'_t\}_t \leftarrow \text{DBSCAN} ( SL_k^{f(a_k)} \cup \{a_k\}, \epsilon, \text{minPts}=2)$ \;
  
  \STATE Update $\mathcal{C}_J$ and $\mathcal{C}_{NC}$ with $\{C'_t\}_t$  \;
  \STATE Update $\mathcal{C}_{U}$ whith clusters from $\mathcal{C}_{NC}$ that are not growing anymore \;
  \STATE Update $n_J$, $n_U$ and $n_{NC}$ \;
 \ENDWHILE
 \RETURN $n_J$, $n_U$ \;
\end{algorithmic}
\end{algorithm}

\subsection{Metrics}

Using these results, we evaluate the proportion of generated UCF with the~2 criteria:

\begin{equation*}
  S_x=\mathds{1}_{n_U>0} \,\,\, \text{ and } \,\,\, R_x= \frac{n_{U}}{n_{U} + n_{J}} \, .
  \label{eq:Bsp_OhmsLaw}
\end{equation*}

We are mainly interested in $S_x$, which labels the studied instance $x$ as being vulnerable to the risk (i.e. having a non-null risk).
The risk itself, measured by $R_x$, describes the likelihood of having an unjustified counterfactual example in the studied area when looking for counterfactual examples.

As these scores are calculated for a specific instance $x$ with a random generation component, we are also interested in their average values $\Bar{S}$ and $\Bar{R}$ for 10 runs of the procedure for each $x$ and over multiple instances.

\subsection{Illustrative Results}
\label{sec:detection-illustrative-results}
\begin{figure}[t]
\centering   
\subfigure{\includegraphics[width=0.47\columnwidth]{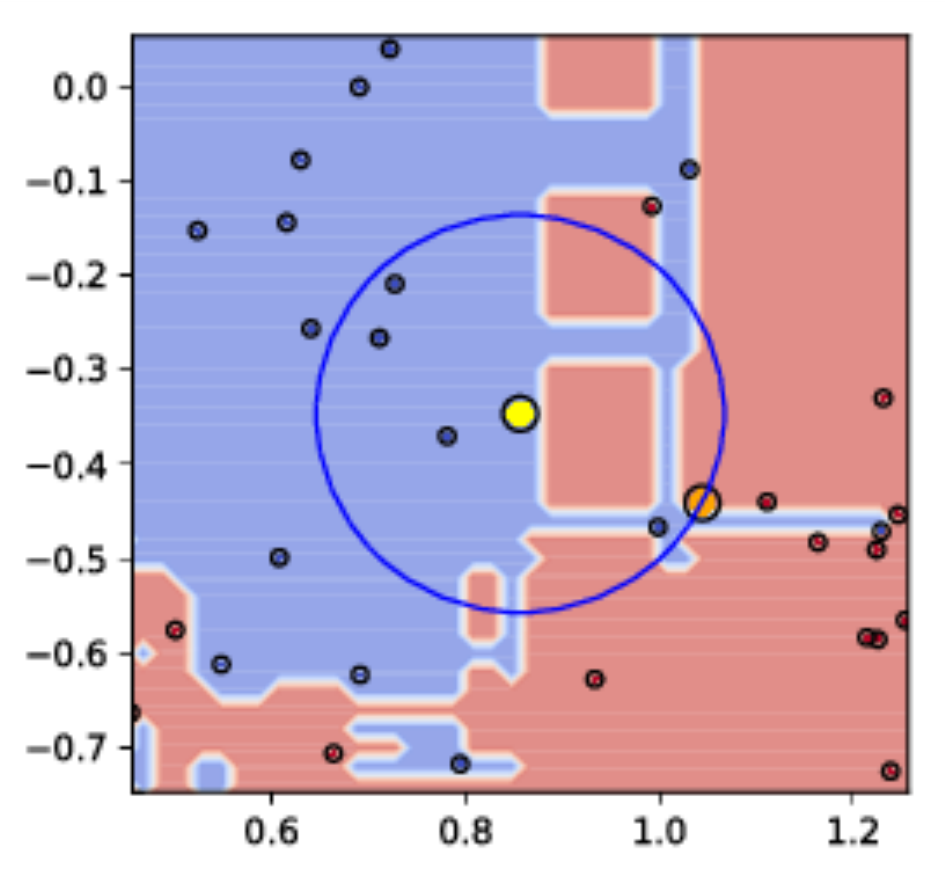}
}
\medskip
\subfigure{\includegraphics[width=0.47\columnwidth]{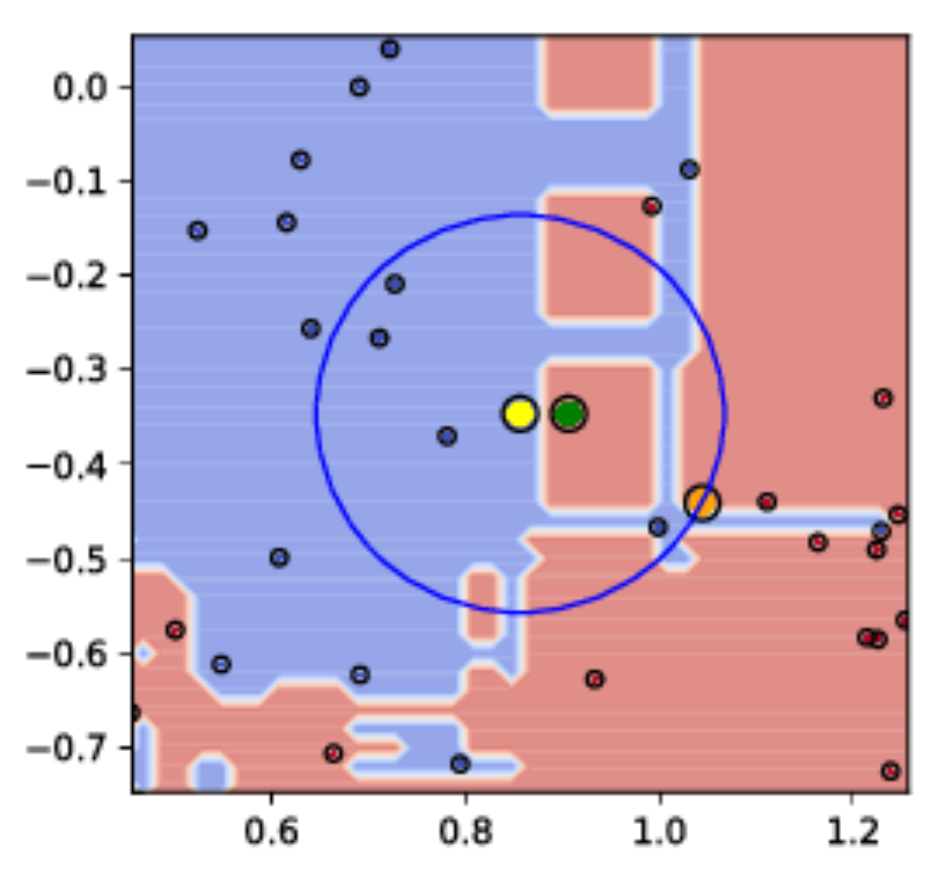}
}
\vspace{-0.3cm}
\caption{Illustrative result of the Local Risk Assessment procedure (left: $S_x=1$) and the Vulnerability Evaluation test (right: $J_x=1$) for an instance of the half-moons dataset.}
\label{fig:illustration-algo-iris}
\vspace{-0.2cm}
\end{figure}

For the purpose of giving insights about the procedure results, a toy dataset (half-moons dataset) is used. A classifier, deliberately chosen for its low complexity (random forest classifiers with only 3 trees), is trained on 70\% of the data (98\% accuracy on the rest of the dataset). The left image of Fig.~\ref{fig:illustration-algo-iris} shows the result of the LRA procedure for a specific instance $x$ (yellow instance), exploring the neighborhood $\mathcal{B}(x, d(x, a_0))$ (blue circle) delimited by its closest neighbor from opposite class $a_0$ (orange instance). The red and blue dots are the instances used to train the classifier, while the red and blue areas represent its decision function: $x$ is predicted to belong to the blue class and $a_0$ to the red one. Within $\mathcal{B}(x, d(x, a_0))$, a red square "pocket" is detected as a group of unjustified counterfactual examples since there is no red instance in this pocket: $S_x = 1$.

This simple example illustrates the fact that the risk of generating unjustified counterfactual examples does exist. Full results of the LRA procedure run over multiple instances of several datasets are available in Section~\ref{sec:results}.

\subsection{LRA Parameters: \texorpdfstring{$n$}{n} and \texorpdfstring{$\epsilon$}{epsilon}}
\label{sec:epsilon}

The values of $n$ and $\epsilon$ are obviously crucial since they define the notion of $\epsilon$-justification and impact the average distance between the number of generated instances $n$. 
Broadly speaking, the higher $n$ and the smaller $\epsilon$ are, the better the approximation of the topology of the local neighborhood is: if the value of $n$ is too low, small "pockets" of unjustified counterfactual examples might be missed. Similarly, if the value of $\epsilon$ is too high, an $\epsilon$-chain will not be a good approximation of a connected path between two instances.
However, there is obviously a tradeoff between this accuracy and the computation time of the algorithm, the complexity of which can be evaluated to be at worst $O(pn^2)$, with $p$ the size of $X^{f(a_0)}$. 
In practice, because the instances~$B_x$ are generated before running DBSCAN, we can set it to the maximum value of the distances of $B$ to their closest neighbors: $\epsilon = \underset{x_i \in B_x}{\max} \underset{x_j \in B_x \setminus \{x_i\}}{\min} d(x_i, x_j)$. However, this requires the whole procedure to be run several time to make sure $\epsilon$ does not take an absurd value (e.g. in the case where one single instance would be generated far away from the others).
Thus, the problem becomes of setting the value of $n$ alone.

The role of parameter $n$ is to make sure that the explored area is "saturated" enough and that no subtelty of the model's decion border, as well as potential unjustified counterfactual examples, are left undetected. Thus, for each observation $x$ we assume there exists a threshold above which the complexity of the decision boundary of $f$ would be fully "captured". When this value is met, increasing $n$ has very little impact over the found clusters and therefore on the risk score $R_x$. We validate this intuition empirically by looking at the evolution of $R_x$ depending on $n$ for multiple instances, but due to space, do not include the results in the present paper. However, the obtained results and code to reproduce them are available in an online repository (\url{https://github.com/thibaultlaugel/truce}). Although seemingly high, the complexity of the LRA procedure is not seen as an issue as it is proposed as a diagnostic tool, rather than used to generate explanations by a user.

\subsection{Quantitative Results}
\label{sec:results}

This section describes the datasets and approaches used in the experiments. 
For each considered dataset, a train-test split of the data is performed with 70\%-30\% proportion, and a binary classifier is trained. To mitigate the impact the choice of the classifier would make, we use the same classifier for every dataset, a random forest (RF) with 200 trees. However, we also train a Support Vector classifier (SVC) with Gaussian kernel on one of the Boston dataset (see below) to make sure the highlighted issue is not a characteristic feature of random forests.

The test set is then used to run the experiments: for each instance it contains, Algorithm~\ref{alg:lra} is run and scores $\Bar{R}$ and $\Bar{S}$ are calculated.
The code and data to reproduce all the experiments are available online in the aforementioned repository.

\subsubsection{Datasets}
The datasets considered for these experiments
include 2 low-dimensional datasets (half-moons and iris) as well as 2 real datasets: Boston Housing~\cite{BostonData} and Propublica Recidivism~\cite{RecidivismData}. 
These structured datasets are commonly used in the interpretability (and fairness) literature.

\subsubsection{Results}
The results of the Local Risk Assessment procedure are shown in Table~\ref{table:detection}.
Among all the considered datasets, between 29\% and 81\% of the studied instances have unjustified counterfactual examples in their neighborhood ($S_x$=1). 

An important thing to note is that the $\Bar{R}$ score varies greatly between the instances of a single dataset (high standard deviation), as well as between the datasets. This can be explained by the local complexity of both the used datasets and classifiers. Supposedly, an instance located far away from the decision boundary of the classifier will have greater chance to generate unjustified counterfactual examples than an instance located closer, since the neighborhood explored by the Local Risk Assessment procedure will be much wider.
More generally, these results depend on assumptions such as characteristics of the classifier (SVC with Gaussian kernel seems to achieve better results than RF), of the data (dimensionality, density) as well as its labels (number of classes, classes separability), and the classifier's accuracy. These have not been studied here as they lie out of the frame of this work, but are somehow illustrated by the variability of $\Bar{R}$ between datasets. The learning algorithm of the classifier itself may also have an impact, as models such as logistic regression or 1-NN classifier have, by design, no UCF ($\Bar{S}=0.0$). Thus, the link between connectedness and overfitting is not obvious and needs to be further explored.

These results confirm the intuition that the risk of generating UCF exists and is important.
In this context, since they do not have the ability to distinguish a ground-truth backed decision from a classifier artifact, post-hoc counterfactual approaches are vulnerable to this issue. The next section studies the extent of this vulnerability.

\begin{table}[t]
  \centering
  \begin{tabular}{lrr} 
     \toprule
 \textbf{Dataset} & $\Bar{S}$ & $\Bar{R}$ [\%] \\
 \midrule
 \textbf{Iris} & $0.41$ & $0.05\,(0.04)$ \\
 \textbf{Half-moons} & $0.37$ & $7.01\,(17.1)$\\
 \textbf{BostonRF} & $0.63$ & $16.0\,(25.2)$ \\
 \textbf{BostonSVC} & $0.29$ & $6.1\,(13.6)$\\ 
 \textbf{Recidivism} & $0.81$ & $26.6\,(30.8)$\\
 \bottomrule
    \end{tabular}
  \caption{Proportion of instances being at risk (average $S$ score) over the test sets, and risk (average percentage and standard deviation values of $R$) of generating an UCF for 5 datasets.}
  \label{table:detection}
  \vspace{-0.1cm}
\end{table}

\section{Vulnerability of Post-hoc CF Approaches}
\label{sec:evaluation}

Once the risk of generating UCF has been established, we analyze how troublesome it is for existing counterfactual approaches. This section presents a second procedure, called \textit{Vulnerability Evaluation}, which aims at assessing how a post-hoc interpretability method behaves in the presence of~UCF.

\subsection{Vulnerability Evaluation Procedure: VE}

The goal of the VE procedure is to assess the risk for state of the art methods to generate UCF in vulnerable regions. 
Given an instance $x\in X$, we use the LRA procedure to assess the risk~$R_x$ and focus on the instances where this risk is "significant" (imposing~$R_x > 0.25$). Using the method to be evaluated, a counterfactual explanation $E(x) \in \mathcal{X}$ is generated.

To check whether $E(x)$ is justified or not, a similar procedure as LRA is used: instances $B_{E(x)}$ are generated uniformly in a local region defined as the hyperball with center~$E(x)$ and radius $d(E(x), b_0)$, where $b_0$ denotes the closest instance to $E(x)$ correctly predicted to belong to the same class, i.e. $b_0 = \underset{x_i \in X^{f(E(x))}}{\text{argmin}} d(E(x),x_i)$. The DBSCAN algorithm is also used with the same parameters as previously. If~$E(x)$ is assigned to the same cluster as the closest instance to~$b_0$, then there exists an $\epsilon$-chain linking $E(x)$ and $b_0$, meaning $E(x)$ is a JCF according to Definition~\ref{def:epsilon-justification}.

If not, similarly as previously, the explored area is expanded to the hyperspherical layer defined by the distance to~$b_1$, the second closest instance from $X^{f(E(x)}$ correctly predicted. This step is repeated by widening the studied area as many times as necessary: if no instance from $X^{f(E(x))}$ can be connected, then $E(x)$ is labelled as being unjustified. 

As for state of the art methods, we focus on the approaches: HCLS~\cite{Lash2017}, Growing Spheres~\cite{Laugel2017inverse} and LORE~\cite{Guidotti2018lore} (modified to return counterfactual examples rather than rules).

\subsection{Metrics}

The goal being to check wether a counterfactual example $E(x)$ is justified or not, we simply define the justification score $J_{E(x)}$ as a binary score that equals $1$ if $E(x)$ is justified, $0$ in the other case.
Again, we also measure the average value $\Bar{J}$ of $J_{E(x)}$ over multiple instances and multiple runs. 

\subsection{Illustrative Results}

The considered post-hoc counterfactual approaches are applied to a toy dataset to illustrate how they can behave, in the aforementioned setup (see Section~\ref{sec:detection-illustrative-results}), see Fig.~\ref{fig:illustration-algo-iris} (right image). As this instance has a risk score $R_x\geq0.25$, we apply VE.
A counterfactual example is generated using HCLS~\cite{Lash2017} with budget $B=d(x, b_0)$. The output $E(x)$ is represented by the green instance. In this situation, $E(x)$ lies in the unjustified area, meaning HCLS generated an unjustified counterfactual example ($J_{E(x)} = 0$). This example illustrates the fact that, when there is a consequent risk $R_x$, post-hoc approaches are vulnerable to it.

\subsection{Quantitative Results}

The results of the VE procedure are shown in Table~\ref{table:evaluation}.
As expected, when confronted to situations with a local risk of generating unjustified counterfactual examples ($R_x\geq 0.25$), the considered approaches fail to generate JCF: the proportion of generated JCF with state of the art approaches can fall as low as 63 \%. This confirms the assumption that post-hoc counterfactual approaches are, by construction, vulnerable to the studied issue.
Among the considered approaches, HCLS seems to achieve slightly better performance than GS. This could be explained by the fact that HCLS tries to maximize the classification confidence, leading to counterfactuals that may be located further from $x$ than GS and hence with a higher chance of avoiding unjustified classification regions. The same observation and conclusion can be made for LORE, which minimizes a $L_0$ norm. Hence, connectedness seems to be able to be favored at the expense of the distance.

Again, a major variability in $\Bar{J}$ can be observed between instances. This confirms the previously made assumption that the connectedness of the classes predicted by~$f$ heavily depends on the data and considered classifier.

\begin{table}[t]
  \centering
  \begin{tabular}{lrrr} 
     \toprule
        \textbf{Dataset} &  \textbf{HCLS} & \textbf{GS} & \textbf{LORE} \\
     \midrule
     \textbf{Iris} & $0.63$ & $0.63$ & $0.70$ \\
     \textbf{Half-moons} & $0.83$ & $0.67$ & $0.83$ \\
     \textbf{BostonSVC} & $0.95$ & $0.93$ & $1.0$ \\
     \textbf{BostonRF} & $0.86$ & $0.84$ & $1.0$\\  
     \textbf{Recidivism} & $0.94$ & $0.94$ & $0.98$ \\
 
 \bottomrule
    \end{tabular}
  \caption{Proportion of generated counterfactuals that are justified ($\Bar{J}$) for vulnerable instances ($R_x \geq 0.25$)}
  \label{table:evaluation}
  \vspace{-0.3cm}
\end{table}

\section{Conclusions}

In this work, we propose a justification definition that counterfactual explanations should satisfy, motivated by an intuitive requirement based on path connectedness. We propose a procedure to assess the risk of generating unjustified counterfactual examples in the post-hoc context, and additionally show that post-hoc counterfactual explanations are indeed vulnerable to such risk.

In this context, we argue that there is no existing satisfying way to provide post-hoc explanations that are both faithful to the classifier and to ground-truth data and further research is necessary. In the mean time, using the train instances, although not always possible in the utilisation context of post-hoc methods, seems to remain necessary.

In light of this study, further works include adapting the current procedure to high-dimensional data and making a more in-depth study of the role of the classifier's complexity in justification. Prospective works also include studying the connectedness of adversarial examples, as ongoing experiments suggest that they verify the proposed justification definition.

\section*{Aknowledgements}
This work has been done as part of the Joint Research Initiative (JRI) project ”Interpretability for human-friendly machine learning models” funded by the AXA Research Fund. We would like to thank Arthur Guillon for stimulating discussions and useful comments.

\clearpage

\bibliographystyle{named}
\bibliography{mendeley}
\end{document}